\documentclass[10pt,twocolumn,letterpaper]{article}

\usepackage{cvpr}      

\usepackage[dvipsnames]{xcolor}

\usepackage{graphicx}
\usepackage{balance}
\usepackage{amsmath}
\DeclareMathOperator*{\argmax}{arg\,max}

\usepackage{amssymb}
\usepackage{booktabs}
\usepackage{multirow, multicol}

\usepackage{algorithm}
\usepackage[noend]{algpseudocode}

\definecolor{cvprblue}{rgb}{0.21,0.49,0.74}
\usepackage[pagebackref,breaklinks,colorlinks,citecolor=cvprblue]{hyperref}

\title{SO-NeRF: Active View Planning for NeRF using Surrogate Objectives}

\author{Keifer Lee$^{1}$,\, Shubham Gupta$^{1}$, Sunglyoung Kim$^{1,2}$, Bhargav Makwana$^{1,2}$,\\ Chao Chen$^{1}$,\, Chen Feng$^{1,}$\thanks{The work is supported by NSF Awards 2238968 and 2322242. The corresponding author is Chen Feng {\tt\small cfeng@nyu.edu}}\\
$^{1}$New York University  $^{2}$Denotes Equal Contribution\\
{\tt\small \url{https://ai4ce.github.io/SO-NeRF}}
}
\begin{document}
\maketitle
\begin{abstract}
\label{sec:abstract}
Despite the great success of Neural Radiance Fields (NeRF), its data-gathering process remains vague with only a general rule of thumb of ``sampling as densely as possible''. The lack of understanding of what actually constitutes good views for NeRF makes it difficult to actively plan a sequence of views that yield the maximal reconstruction quality. We propose Surrogate Objectives for Active Radiance Fields (SOAR), which is a set of interpretable functions that evaluates the goodness of views using geometric and photometric visual cues - surface coverage, geometric complexity, textural complexity, and ray diversity. Moreover, by learning to infer the SOAR scores from a deep network, SOARNet, we are able to effectively select views in mere seconds instead of hours, without the need for prior visits to all the candidate views or training any radiance field during such planning. Our experiments show SOARNet outperforms the baselines with $\sim$80x speed-up while achieving better or comparable reconstruction qualities. We finally show that SOAR is model-agnostic, thus it generalizes across fully neural-implicit to fully explicit approaches.
\end{abstract}    
\section{Introduction}
\label{sec:intro}

A key aspect of a practical and intelligent embodied agent is the ability to dynamically respond and plan appropriately to an ever-changing environment - dubbed active perception \cite{aperception}. Of course, one presupposes the ability to construct a reliable representation of the environment as a prior, and a class of approach has been found to be particularly efficacious in recent years - neural implicit representations with Neural Radiance Fields (NeRF) \cite{mildenhall2020nerf, arandjelovic2021nerf}. Its applications ranges from Virtual Reality \cite{VRNeRF, deng2022fov, immersive-nerf} to Simultaneous Localization, Mapping (SLAM) \cite{Zhu2022CVPR, rosinol2022nerf, h2-nerfslam}. 

However, in contrast to their popularity, the prototypical process of constructing such representations remained vague with a general rule of thumb - capture as many images of the subject as densely and directionally varied as possible. The latter suggestion may be fine for one-off tasks or small subjects, but grows to be intractable when either of the aforementioned conditions are not satisfied; e.g. capturing images for thousands of products or trying to reconstruct the Santa Maria del Fiore in its entirety for large-scale exploration and neural topological mapping. In robotics context, a properly perceived representation is key to planning and navigation. Therefore, it is of interest to study more efficient ways of capturing the required images, and indeed this is known as the Next Best View (NBV) problem \cite{nbv-ori}.  

There have been many proposed approaches to tackle the NBV problem in the classical Structure from Motion (SfM) setting \cite{guedon2022scone, guedon2023macarons}, and even in the recent neural implicit framework \cite{pan2022activenerf, jin2023neu, goli2023bayes, neurar, jain2021putting, yu2021pixelnerf, chen2021mvsnerf}. However, these methods so far suffer from some key problems from an application-viability perspective in one way or another. Probabilistic uncertainty estimation methods such as ActiveNeRF \cite{pan2022activenerf}, NeurAR \cite{neurar} and Bayes' Rays \cite{goli2023bayes} are either too slow or not straightforward to work with by requiring training of a NeRF model - effectively an in-flight or post hoc approach; they are also not interpretable as abstract probabilistic uncertainties does not correspond to actual physical qualities. On the other end of the spectrum, there are methods from classical SfM tasks such as SCONE \cite{guedon2022scone} that utilizes intuitive concepts like surface coverage as a proxy objective to maximize for. However, these uni-dimensional methods are too simplistic, which may not generalize well across object classes with different structure.

Therefore, in this work, we introduce SOAR to tackle the aforementioned problems of active perception for radiance fields models such as NeRF. Specifically, the main contributions of our work are as follows: 
\begin{enumerate}
  \item We propose a set of surrogate objectives, SOAR to characterize informative samples for training radiance fields models. These functions correspond directly to physical qualities, 
  rendering them intuitive and explainable.
  \item Unlike in-flight and post hoc methods, SOAR is prescriptive, not needing any radiance field model training at all. Therefore it is also radiance fields model-agnostic.
  \item We further devised a method to efficiently compute these objectives with a novel neural network, dubbed SOARNet, generating trajectories in seconds instead of hours.
  \item Finally, SOAR provides an alternative training-free approach to estimating the goodness of a radiance field model with respect to the input data. We will release our code and data for reproducibility.
\end{enumerate}
\label{sec:literature}
\begin{figure*}[t]
  \centering
    \includegraphics[width=1\linewidth]
    {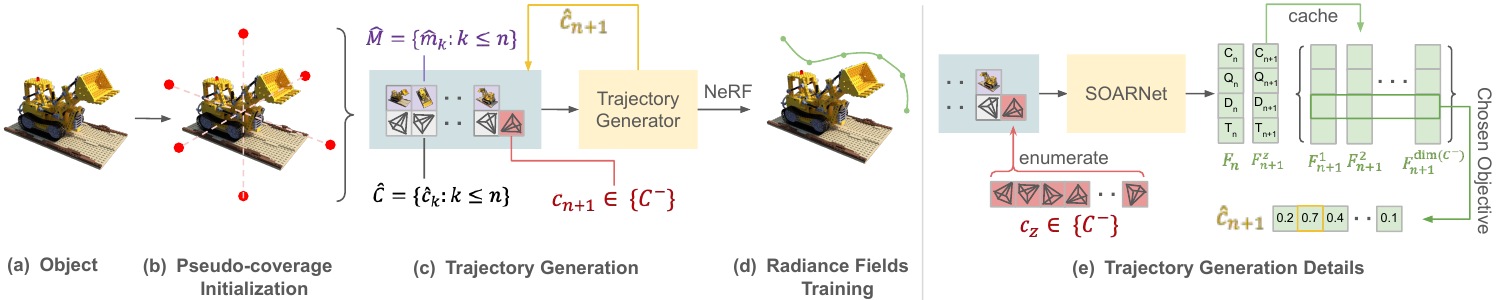}
    \caption{The overall pipeline of our proposed approach. \textbf{(a)} Given an object, \textbf{(b)} we initialize via pseudo-coverage initialization to obtain a set of starting poses $\mathcal{\hat{M}}$ and corresponding camera-poses $\mathcal{\hat{C}}$. \textbf{(c)} We then generate a trajectory for sampling in a greedy manner by computing the objective scores of all unseen camera-poses, $c_{n+1}$ where $\mathcal{C}^-$ is the set of all unseen camera-poses. \textbf{(e)} To compute said scores, we utilize our novel model, SOARNet for speedy inference, and we then simply pick the candidate with the highest score, $c_z$ as our next pose, $\hat{c}_{n+1}$ until we reach our budget, $\mathcal{B}$ for $\textbf{(d)}$ radiance fields training.}
    \label{fig:cqdt_architecture}
\end{figure*}

\section{Related Work}

\noindent{}\textbf{Novel View Synthesis}
is generating views of the given object or scene from previously unseen camera poses. This is a challenging task due to the high uncertainty of elements that were previously occluded from the field of view \cite{Yu2023PhotoconsistentNVS, liu2023robust, liu2022neural, Zhu_2023_CVPR}. While there are many approaches recently published \cite{chan2023generative, kerbl3Dgaussians, wynn-2023-diffusionerf, zeng2020pc}, we only explore Novel View Synthesis using radiance fields \cite{mildenhall2020nerf, gao2022nerf} to stay within the scope of this paper. Note that the original NeRF \cite{mildenhall2020nerf} by Mildenhall et al. shall be referenced as \textit{Vanilla NeRF} interchangeably henceforth to distinct the generic idea of neural radiance field modeling and the actual model itself.

The original NeRF while impressive was not very computationally efficient as it encodes the object via a neural network (implicit representation). To further improve the run-time performance, there have been recent works alleviating the computation bottleneck \cite{garbin2021fastnerf}, and these works can be categorized as baked (fully-implicit), hybrid, and fully-explicit representation methods \cite{hedman2021snerg, mueller2022instant, plenoxels, Chen2022ECCV}. Fridovich-Keil et al. \cite{plenoxels} voxelize the scene and store the density and spherical harmonics coefficients as scalars and vectors per color channel respectively. As such, slow neural networks are not required, resulting in what's known as a fully-explicit representation. M\"{u}ller et al. \cite{mueller2022instant} leveraged a multi-resolution hash table of feature vectors to lookup the pixel's features, resulting in a massive speed-up. Since their approach leverages a neural network for volumetric rendering, such methods are classified as hybrid representation approaches. In the same class, Chen et al. \cite{Chen2022ECCV} proposed a method with massive speed-up and saving in memory footprint compared to fully-explicit approaches by representing a scene as a 4D tensor and decomposing the tensor into multiple low-rank components. 
\\

\noindent{}\textbf{Next Best View} planning is a highly relevant topic for robotic perception that deals with finding the best sensor positioning to maximize information gain. This also directly ties in with the idea of active perception \cite{aperception, bajcsy2018revisiting}, where the observer moves around a scene with an intention to better capture salient information. Active perception has previously been employed for various tasks \cite{ren2019domain, novkovic2020object, wang2020v2vnet} but only recently for training radiance fields model \cite{pan2022activenerf}.

Gu\'{e}don et al. \cite{guedon2022scone} propose a hybrid method that predicts an occupancy grid so as to maximize the total surface coverage using a neural networks and Monte Carlo integration over a volumetric representation. Improving on that, \cite{guedon2023macarons} utilizes only camera inputs and is able learn in a self-supervised fashion. Kopanas and Drettakis \cite{10.2312:vmv.20231222} introduced observation frequency and angular uniformity as metrics to improve convergence for free-viewpoint navigation. Dhami et al. \cite{dhami2023pred} propose a two-stage process that predicts the complete point cloud using partial observations and selects for views that minimizes the distance traveled.
\\
\noindent{}\textbf{Uncertainty estimation in NeRF} has recently received a lot of interest. Pan et al. \cite{pan2022activenerf} incorporates uncertainty estimation into the Vanilla NeRF training pipeline, which learns to estimate the uncertainty of candidate poses in parallel during training via Bayesian estimation. These candidates are then greedily selected post-training to maximize the estimated information gain. Lee et al. \cite{lee2022uncertainty} proposes a ray-based uncertainty estimator by computing the entropy of the distribution of the color samples along each ray. Similarly, Shen et al. \cite{shen2021stochastic} presents a generalized NeRF model that can learn the probability distribution of all radiance fields representing a given scene for uncertainty estimation via Bayesian learning. Jin et al. \cite{jin2023neu} proposes a generic framework that learns to predict the uncertainty of each candidate poses, but without needing to train a radiance fields model. However, to compute the uncertainties, their method still requires the images taken from said candidate poses in the first place, which is a non-negligible limitation in real-world applications.

Goli et al.  \cite{goli2023bayes} presents a post hoc framework to evaluate the uncertainty in any pretrained NeRF model. The framework produces a spatial uncertainty field by introducing perturbations along the learned camera rays and by utilizing Bayesian Laplace approximations. Ran et al. \cite{neurar} proposes a proxy for the Peak Signal-to-Noise Ratio (PSNR) metric that is jointly trained with the parameters of an implicit radiance fields model to predict a new quantity, dubbed Neural Uncertainty for trajectory planning. Our approach, does not perform post-hoc uncertainty estimation, but preemptively evaluates unseen candidate poses via the SOAR scores.
\section{Methodology}
We begin by formulating the problem - given an object and a corresponding set of images of said object, $\mathcal{M}=\left\{m_1,m_2,..., m_N\right\}$, taken from camera-poses, $\mathcal{C}=\left\{c_1,c_2,..., c_N\right\}$, $c \in \mathbb{R}^{3\text{x}4}$ and a budget constraint, $\mathcal{B} \in \mathbb{N}$ we want to select a subset of poses, subjected to the constraint, that will maximize a reconstruction quality metric such as PSNR, $\Gamma$ via a radiance fields model, $\Omega$. Formally,
\begin{equation}
  \hat{\mathcal{C}} = \argmax_{c} \Gamma(\Omega(\mathcal{C},\mathcal{M})) \quad s.t. \;\, \text{dim}(\hat{\mathcal{C}})=\mathcal{B}
  \label{eq:problem_def}
\end{equation}
In the following subsections, we outline the different parts of the pipeline. In \ref{subsec:method_defining_obj_funcs}, we formally define the objective functions of our approach from first principles. In \ref{subsec:method_cqdt_model}, we introduce SOARNet which enables us to efficiently compute the objective scores. Lastly, in \ref{subsec:method_trajectory_generation} we introduce the finalized framework by delineating a simple trajectory planning schema with SOARNet.

\subsection{Objective Functions Definitions}
\label{subsec:method_defining_obj_funcs}
We begin with a set of presuppositions on the qualities of a training set that would yield a NeRF with good reconstruction as measured via $\Gamma$. Fundamentally, we hypothesized from first principles that a good training set can be obtained by maximizing the surface coverage of the object, the attention paid to local regions with particular geometric or textural complexity, and the resulting multi-view diversity of observations. Succinctly, we refer to the aforementioned qualities as the (i) Surface Coverage, ${\mathbf{C}}$ (ii) Geometric Complexity, $\mathbf{Q}$ (iii) Textural Complexity, $\mathbf{T}$ (iv) and Ray Diversity, $\mathbf{D}$ respectively. 
\\

\noindent\textbf{Surface Coverage}, $\mathbf{C}$ as defined is simply the ratio of seen faces, ${j}$ via the image taken from the $k$-th camera-pose, ${c_k}$, against the total number of faces of the object or mesh, ${J}$. We can therefore define a corresponding objective function for $\mathbf{C}$ as follows, where $f_C \in \mathbb{R}$
\begin{equation}
  f_C = \frac{1}{J} \sum_{j}^{J} \mathrm{U}(\sum_{k}^{n} V_{j}^{k})
  \label{eq:obj_C}
\end{equation}
\noindent{}where $\mathrm{U(\cdot)}$ is the Heaviside step function, $n$ is the total number of camera-poses visited, and $V_{j}^{k}$ is an indicator function corresponding to the visibility of face $j$ from the $k$-th camera-pose. Given the mesh of an object, optimizing for Eq. (\ref{eq:obj_C}) amounts to maximizing the number of unique faces seen across the camera-poses.
\\ 

\noindent\textbf{Geometric Complexity}, $\mathbf{Q}$ corresponds to the variability of the mesh faces' configurations, and one way to quantify this complexity is via the mesh normals, $\hat{\mathrm{n}}$. For example, a geometrically complex mesh such as a cathedral, would have a higher degree of surface normal variability compared to a simple smooth ball due to the varied substructures comprising the cathedral - e.g. pillars, the dome. In the case of the cathedral, we would intuitively want to pay extra attention to the local regions that are complex by sampling more varied views of said regions.

We formally define the objective function for $\mathbf{Q}$ as follows, where $f_Q \in \mathbb{R}$
\begin{equation}
  f_Q' = \left|\mathrm{LoG}(\hat{\mathrm{n}}_j; \sigma_Q)\right| \, \beta_Q \, V_{j}^{k} + 1
  \label{eq:obj_Q_1}
\end{equation}
\begin{equation}
  f_Q = \frac{1}{J(5\sigma_Q)^2} \sum_{j}^{J} f_Q'
  \label{eq:obj_Q_2}
\end{equation}
\noindent{}where $\mathrm{LoG(\cdot)}$ is the Laplacian of Gaussian filter, $\sigma_Q$ is the standard deviation of the Gaussian filter, $\beta_Q$ a scaling factor, $f_Q'$ is the per-face geometric complexity and $f_Q$ is the normalized total complexity. We take the absolute value of the output since the directionality of the normal variations does not matter. Maximization of Eq. {\ref{eq:obj_Q_2}} will encourage the selection of camera-poses with higher geometric complexity as defined above. In practice, a scaling discount factor, $d_Q$ is applied towards the $\mathbf{Q}$ values of overlapping faces seen during trajectory planning in order to control for over-emphasis on certain local regions with unusually high geometric complexity, trapping it in a local maxima.
\\

\noindent\textbf{Textural Complexity}, $\mathbf{T}$ corresponds to the photometric variation of images taken at camera-pose, $c_k$ of the object. This means that $\mathbf{T}$ is computed with respect to the 2D projections of the object instead of the underlying 3D mesh. We have chosen this simplified approach for $\mathbf{T}$ since unlike geometric complexities which may be hard to discern on 2D projections, textural characteristics are not. Similar to Geometric Complexity, $\mathbf{Q}$ we hypothesized that regions with high textural complexity should be paid extra attention. Formally, we define the corresponding objective function for $\mathbf{T}$ as follows, where $f_T \in \mathbb{R}$
\begin{equation}
  f_T' = \mathrm{Hist}\left( \mathrm{LBP} \left(m_k; r_T, p_T\right) \right)
  \label{eq:obj_T_1}
\end{equation}
\begin{equation}
  f_T = 1 - f_T'\left[b\right]
  \label{eq:obj_T_2}
\end{equation}
\noindent{}where $\mathrm{Hist(\cdot)}$ is simply the histogram function, $b$ the last bin's index, $\mathrm{LBP(\cdot)}$ is the Local Binary Pattern texture descriptor \cite{LBP}, $r_T$ and $p_T$ are the filter radius and the number of circularly symmetric local neighborhood points of the descriptor, and finally $m_k$ is the image taken at camera-pose, $c_k$. Similar to $f_Q$, in practice, we also impose a discount factor, $d_T$ towards the $\mathbf{T}$ values seen during trajectory planning to prevent overfitting, and we used $r_T$ of 3 and $p_T$ of 3. Finally, we compute the final score by averaging $f_T$ computed on the Hue and Saturation channels of the image respectively. Maximization of Eq. (\ref{eq:obj_T_2}) is equivalent to selecting the image with the highest textural complexity. 
\\

\begin{figure*}[t]
  \centering
    \includegraphics[width=0.85\linewidth]
    {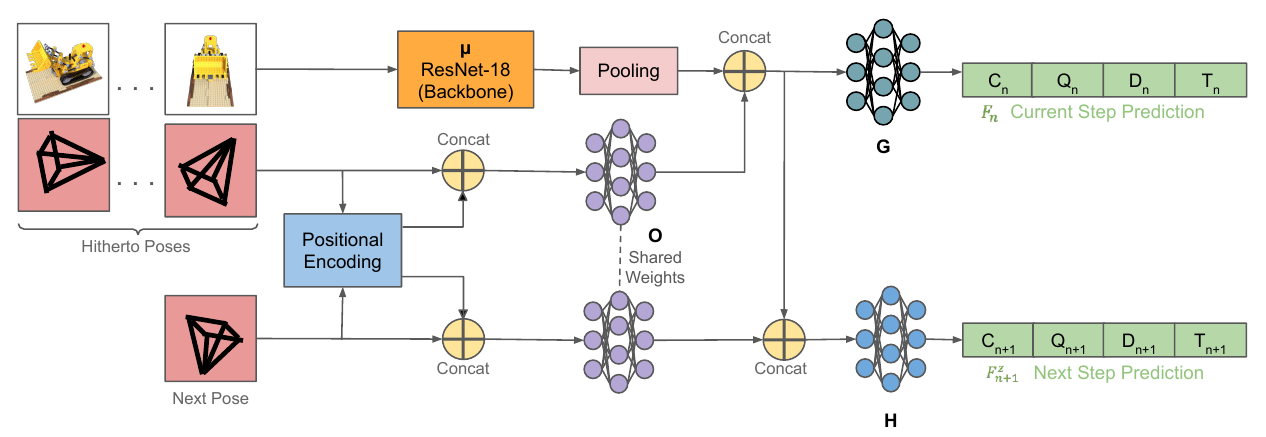}
    \vspace{-2mm}
    \caption{Architecture of the SOARNet.}
    \vspace{-2mm}
    \label{fig:cqdt_model}
\end{figure*}

\noindent\textbf{Ray Diversity}, $\mathbf{D}$ as defined is the variation of the rays, $i$ projected from various viewing poses observing a particular face, $j$ of the object mesh. Intuitively, having highly varied novel views of the object faces would yield better reconstruction. The idea is closely related to densely sampling a set of training images from as many views as possible to improve multi-view correspondences. We went a step further here by exactly quantifying said idea with a formal definition of the objective function as follows
\begin{equation}
   \theta^k_{ij} = \left| \frac{180^\circ}{\pi} \arccos{\left(\frac{i\hat{\mathrm{n}}_j}{\left| i \right| \left|\hat{\mathrm{n}}_j \right|}\right)} \right| - 90^\circ
  \label{eq:obj_D_1}
\end{equation}
where $\theta^k_{ij}$ is the augmented ray-surface grazing angle between ray  $i$ from camera-pose $c_k$ and face $j$ with surface normal $\hat{\mathrm{n}}_j$. However, as opposed to indiscriminately maximizing for view-angle variance, we impose a penalty towards any ray observing a face beyond a certain grazing angle via soft-thresholding. The intuition behind the penalty is derived from the observation that it becomes increasingly difficult to visually obtain useful information about a given object's face as we approach the orthogonal angle of its surface normal. We perform said soft-thresholding by assigning each observing ray, $i$ with an outlier score 
\begin{equation}
   \psi^k_{ij} = 1 - \frac{\alpha_1}{1+e^{-\theta^k_{ij}+\alpha_2}}
  \label{eq:obj_D_2}
\end{equation}
where $\psi^k_{ij}$ is the outlier score, and $\alpha_1$ and $\alpha_2$ are hyperparameters to control the grazing angle soft-threshold with values of $1$ and $-10$ respectively. We then compute the total outlier ratio of face $j$ as follows
\begin{equation}
   \phi_j = \frac{1}{K_j} \sum_k^n \left( \sum_i^I \psi^k_{ij} V^k_{ij} \right) 
  \label{eq:obj_D_3}
\end{equation}
\noindent{}where 
\begin{equation}
   K_j = \sum_k^n  \sum_i^I V^k_{ij}
  \label{eq:obj_D_4}
\end{equation}
\noindent{}which simplifies Eq. (\ref{eq:obj_D_3})
\begin{equation}
   \phi_j = \mathbb{E} \left[ \psi^k_{ij}\right]
  \label{eq:obj_D_5}
\end{equation}
Next, we compute the ray diversity, $d_j$ of face $j$ as follows
\begin{equation}
   \delta_j = \frac{1}{I^3} \left[ \sum_{i}^{I} \left( x_i - \Bar{x}_i \right)^2 \sum_{i}^{I} \left( y_i - \Bar{y}_i \right)^2 \sum_{i}^{I} \left( z_i - \Bar{z}_i \right)^2 \right]
  \label{eq:obj_D_6}
\end{equation}
where $x,y,z$ are the corresponding principal axis values of each observing ray $i$. This is akin to multiplying the variances of each individual principal axis of the observing rays.
\begin{equation}
   \delta_j = \sigma_x^2 \sigma_y^2 \sigma_z^2
  \label{eq:obj_D_7}
\end{equation}
Finally, the normalized effective Ray Diversity objective function is as follows, where $f_D \in \mathbb{R}$
\begin{equation}
   f_D = \frac{1}{J} \sum_j^J \delta_j\phi_j
  \label{eq:obj_D_7_2} 
\end{equation}
Note that for all the objective functions of $f_C$, $f_Q$, $f_D$, and $f_T$, their values are further smoothened and clipped with the following smoothing function, $\xi:\mathbb{R} \mapsto \left[0,1\right]$ with $\alpha_3$ of $3$.
\begin{equation}
   \xi(x) = \mathrm{max} \left(\mathrm{min} \left( \frac{1}{1+ \left[ e^{(-\alpha_3 e x) + (\alpha_30.5e)} \right] } ,1 \right) ,0 \right)
  \label{eq:smoothing}
\end{equation}

\subsection{Efficient Computation with SOARNet}
\label{subsec:method_cqdt_model}

With the objective functions defined, we now turn our attention to efficient computation. By default, the ground truth mesh is required to compute the objective functions, which limits its applicability. In order to remediate this, we proposed a Deep Neural Network, \textbf{SOARNet} to infer the respective objective scores for the poses visited so far, $\mathcal{\hat{C}} = \left\{ \hat{c}_k: k \leq n \right\}$ and a further unvisited pose, $c_{n+1}$ given a set of corresponding monocular 2D images up to step $n$, $\mathcal{\hat{M}} = \left\{ m_{k}: k \leq n \right\}$, with constraint $n+1 \leq \mathcal{B}$, and the aforementioned camera parameters of each image; note that our method does not require the image of the unvisited camera-pose as input. Formally,
\begin{equation}
   \textbf{SOARNet}\left( \mathcal{\hat{C}},\mathcal{\hat{M}},c_{n+1} \right) = \left\{ \mathbf{F}_g: g \in \{ n, n+1\} \right\}
  \label{eq:cqdt_eq_1}
\end{equation}
\noindent{}where
\begin{equation}
   \mathbf{F}_g = \left\{f_C^g, f_Q^g, f_D^g, f_T^g \right\}
  \label{eq:cqdt_eq_2}
\end{equation}
At each step, the elements of $\mathbf{F}_n$ correspond to the normalized cumulative scores of all poses seen so far for the corresponding objectives. Likewise, $\mathbf{F}_{n+1}$ returns the normalized cumulative scores from an unseen camera-pose, $c_{n+1}$ conditioned upon the visited poses.

The architecture of the model can be found in Fig. \ref{fig:cqdt_model}, where $\mu(\cdot)$ is a ResNet \cite{resnet} encoder, $o(\cdot)$ is an MLP encoder taking positionally encoded camera extrinsics, and $G(\cdot)$ and $H(\cdot)$ are simple MLP based prediction heads for generating the corresponding objective scores, $\mathbf{F}_g$ via regression. For loss functions, we used the Huber loss on the predictions of each head and take their sum as the final loss, $\mathcal{L}$ as follows
\begin{equation}
   %
   \mathcal{L} = \mathcal{L}_{\text{G}}\left( \mathbf{\hat{F}}_n, \mathbf{F}_n \right) + \mathcal{L}_{\text{H}}\left( \mathbf{\hat{F}}_{n+1}, \mathbf{F}_{n+1} \right)
  \label{eq:loss_fx_1}
\end{equation}


\begin{algorithm}[t]
\caption{Trajectory Planning}\label{algo:trajectory}
\begin{algorithmic}[1]
\Procedure{GenerateTrajectory}{}
    \State $B \gets \text{length of } \text{budget}$
    \State $\mathcal{P} \gets \text{objective sequence, } [\mathbf{C},\mathbf{Q},\mathbf{Q},\mathbf{D},\mathbf{D},\mathbf{T}]$
    \State $\mathcal{C}^{-} \gets \text{set of unseen poses}$
    \State $\mathcal{\hat{C}} \gets \text{camera poses list}$
    \State $\mathcal{\hat{M}} \gets \text{images list}$
    \State $n_{\text{init}} \gets \text{number of initialization steps}$
    \While {step $\leq B$}
        \If {step $\leq n_{\text{init}}$}
            \State \textbf{do} $\textit{PseudoCoverage}$

        \Else{}
            \State $idx \gets step \; \% \; \textit{len}(\mathcal{P})$
            \State objective, $f$ $\gets \mathcal{P} [idx]$
            \State $scores \gets \text{empty } array$
            \For{$c_z$ in $\mathcal{C}^{-}$}
                \State $s_f \xleftarrow[]{\text{select} f} \text{SOARNet}\left(\mathcal{\hat{C}},\mathcal{\hat{M}},c_{z} \right)$
                \State $scores.append\left(s_f\right)$
          \EndFor
          \State $\mathcal{\hat{C}}.append\left( argmax(scores)\right)$  
          \State $update$ $\mathcal{\hat{M}}$
        \EndIf
        \State $step \gets step + 1$
    \EndWhile
    \State \textbf{done}
\EndProcedure
\end{algorithmic}
\end{algorithm}

\subsection{Trajectory Generation with SOARNet}
\label{subsec:method_trajectory_generation}
Once we are able to determine the importance of each candidate pose, the next step would be constructing an optimal trajectory. We propose a simple yet effective greedy strategy by picking the view that maximizes the desired objective at each step. 
At once, another question naturally follows from the proposed approach, which objective function should be optimized for? We observed that the optimal objective function, is class-dependent - some object classes are better addressed by surface coverage, $\mathbf{C}$ and some via textural complexity, $\mathbf{T}$ instead. Intuitively, this makes sense since each object may physically and visually differ significantly. Therefore, if we simply maximize the total objective scores, it may not actually be picking the candidate with the maximal marginal score of the most effective objective function for the object.


Hence, our trajectory planning algorithm maximizes only a single objective per step in an interleaved manner. This way, we are able to generate a trajectory that accounts for all objectives as a whole. From our empirical testing, we have found that $\mathbf{D}$ and $\mathbf{Q}$ to be most helpful, and therefore we gave particular emphasis to these objectives in our ensemble with the following sequence, $[\mathbf{C},\mathbf{Q},\mathbf{Q},\mathbf{D},\mathbf{D},\mathbf{T}]$; for brevity, we refer to this as the ensemble objective, $\mathbf{E}$. 

Finally, we hypothesize and indeed observe that having good coverage is of particular importance during initialization when available information is minimal. As such, we utilize an effective yet easy-to-compute pseudo-coverage initialization schema which simply picks poses at the extremum along each principal spatial axis for $n_{\text{init}}$ number of poses to begin with. Succinctly, the trajectory planning schema for $\mathbf{E}$ is summarized in Algorithm \ref{algo:trajectory} with $n_{\text{init}}=3$.
\section{Experiment}\label{sec:experiment}

In this section, we first present our experimental dataset and setup in Sec. \ref{subsec:dataset_and_setup}, before delving more into the results with respect to various baselines in Sec.\ref{subsec:results}.

\subsection{Dataset and Setup}
\label{subsec:dataset_and_setup}

\begin{table*}[t]
\centering
    \begin{tabular}{lcc|cc|cc|cccl}
    \toprule
        {} & \multicolumn{2}{c}{\textbf{PSNR} ($\uparrow$)} & \multicolumn{2}{c}{\textbf{LPIPS} ($\downarrow$)} & \multicolumn{2}{c}{\textbf{SSIM} ($\uparrow$)} & \multicolumn{4}{c}{\textbf{Time} ($\downarrow$)}\\
        \midrule
        Object & AN & Ours  &    AN & Ours &   AN & Ours  &      AN (TG) &  AN (RFT)  & Ours (TG) & Ours (RFT)   \\
        \midrule
drums     &      17.92 &  \textbf{18.25} &      0.332 &  \textbf{0.322} &      \textbf{0.779} &  0.774 &  \multicolumn{2}{c}{14h 12m}  &  \textbf{20s} & \textbf{11m 42s}\\
ficus     &      21.55 &  \textbf{21.81} &      0.169 &  \textbf{0.163} &      0.869 &  \textbf{0.877} &    \multicolumn{2}{c}{14h 18m}   &  \textbf{20s} & \textbf{15m 6s}\\
hotdog    &      25.36 &  \textbf{28.77} &      0.202 &  \textbf{0.123} &      0.888 &  \textbf{0.921} &    \multicolumn{2}{c}{14h 10m}   &  \textbf{20s} & \textbf{10m 13s}\\
lego      &      23.57 &  \textbf{25.17} &      0.169 &  \textbf{0.127} &      0.835 &  \textbf{0.865} &   \multicolumn{2}{c}{14h 12m}   &  \textbf{20s} & \textbf{9m 46s}\\
materials &      19.54 &  \textbf{21.27} &      0.234 &  \textbf{0.192} &      0.811 &  \textbf{0.839} &   \multicolumn{2}{c}{14h 22m}    &  \textbf{20s} & \textbf{10m 4s}\\
mic       &      24.66 &  \textbf{24.94} &      0.146 &  \textbf{0.135} &      0.913 &  \textbf{0.918} &   \multicolumn{2}{c}{14h 14m}   &  \textbf{20s} & \textbf{8m 36s}\\
ship      &      \textbf{21.27} &  21.14 &      0.334 &  \textbf{0.332} &      \textbf{0.721} &  0.716 &   \multicolumn{2}{c}{14h 17m}   &  \textbf{20s} & \textbf{11m 28s}\\
    \bottomrule
\end{tabular}
  \caption{Reconstruction quality of \textbf{TensoRF} \cite{Chen2022ECCV} trained via trajectory generated \textbf{ActiveNeRF, AN} \cite{pan2022activenerf} and \textbf{Ensemble} (Ours) on NeRF Synthetic. For time, \textbf{TG} is the trajectory generation time, and \textbf{RFT} is the radiance fields training time. Best result is represented with \textbf{bold}.}
  \label{tab:result_2}
\end{table*}



\noindent{}\textbf{Dataset}. For our experiments, we have created a diverse dataset of 640 objects across 20 classes based on the Objaverse \cite{deitke2023objaverse} dataset containing Large Vocabulary Instance Segmentation categories \cite{gupta2019lvis}. Similar to NeRF-Synthetic \cite{mildenhall2020nerf}, for each object we sample images taken from random poses on a fixed-size spherical surface looking inward towards the object, with 100 images for training, 100 for validation and 200 for testing using Blender. 
More information on the generation process can be found in supplementary. 

From this set of objects, we manually computed the individual $f_C$, $f_Q$, $f_D$, and $f_T$ objectives' score, and generated corresponding trajectories of length $30$ via a random walk on the full train-set. Our final train, validation, and test sets have approximately $31$k, $1.1$k, and $5$k samples respectively. For the final evaluation with actual radiance fields training, we further selected a subset of $50$ objects from the test set to do so for computational feasibility; this subset will be referred to as the \textit{final-set} for brevity. Finally, we also leveraged the NeRF-Synthetic dataset for out-of-distribution performance test.
\\

\noindent{}\textbf{Modelling Radiance Fields}. For radiance fields modeling, we utilized TensoRF \cite{Chen2022ECCV} as our primary model. TensoRF adopts a hybrid approach to neural radiance fields modeling unlike the Vanilla NeRF, which enables it to achieve significant speed-up in model training and convergence via tensor decomposition. However, since the proposed objective functions are agnostic with respect to the radiance fields modeling method, we hypothesized that it should be effective regardless of the variants. Indeed, we show that our method generalizes across radiance fields model variants in Sec. \ref{subsec:results}. 
\\

\noindent{}\textbf{Experimental Setup}. The \textbf{SOARNet} model was trained on a machine with an NVIDIA RTX8000 GPU, 64GB of DRAM and $8$ CPUs. The model was trained with a step-scheduler of step size of $0.6$ and $\gamma$ of $0.5$, and the Adam optimizer \cite{kingma2014adam} at a learning rate of $3e-5$, $\beta_1$ of $0.9$ and $\beta_2$ of $0.999$, $\epsilon$ of $1e-08$, with AMSGrad \cite{amsgrad} enabled, and no weight decay. Finally, we trained the model for $12$ epochs with a batch-size of $8$. 

We leverage the NerfStudio \cite{nerfstudio} implementations of TensoRF \cite{Chen2022ECCV} and Vanilla-NeRF \cite{mildenhall2020nerf} for radiance field training. We also use Plenoxels \cite{plenoxels} and ActiveNeRF \cite{pan2022activenerf} provided by the respective authors. For details on the exact configurations, please refer to supplementary.


\subsection{Results Evaluation}
\label{subsec:results}

\begin{figure*}[t]
  \centering
    \includegraphics[width=0.68\textwidth]
    {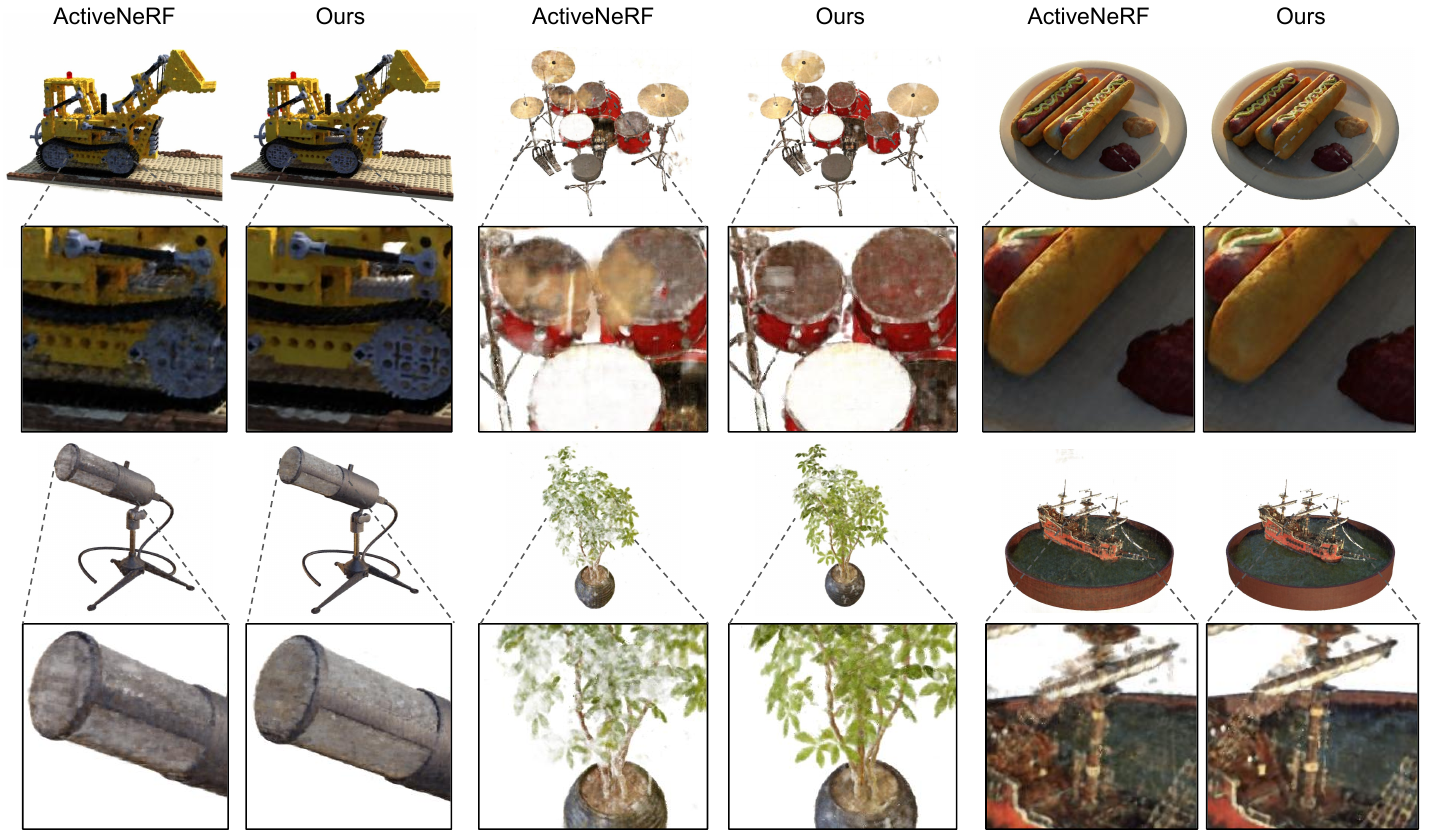}
    \caption{Example renders with our method against the baseline, ActiveNeRF. Here we used a budget of $\mathcal{B}=30$ for all instances.}
    \vspace{-2mm}
    \label{fig:result_render}
\end{figure*}
\begin{table*}[t]
\centering
    \begin{tabular}{lcc|cc|cc||cc|cc|cc}
    \toprule[1.5pt]
        \multicolumn{7}{c}{\textbf{Vanilla NeRF}} & \multicolumn{6}{c}{\textbf{Plenoxels}}  \\
        \midrule[1.5pt]
        {} & \multicolumn{2}{c}{\textbf{PSNR} ($\uparrow$)} & \multicolumn{2}{c}{\textbf{LPIPS} ($\downarrow$)} & \multicolumn{2}{c||}{\textbf{SSIM} ($\uparrow$)} & \multicolumn{2}{c}{\textbf{PSNR} ($\uparrow$)} & \multicolumn{2}{c}{\textbf{LPIPS} ($\downarrow$)} & \multicolumn{2}{c}{\textbf{SSIM} ($\uparrow$)} \\
        \midrule
        Object & Rand & Ours &    Rand & Ours &    Rand &    Ours & Rand & Ours &    Rand & Ours &    Rand & Ours \\
        \midrule
            bookcase  &  33.83 & \textbf{35.02} &  0.042 & \textbf{0.038} &  0.975 & \textbf{0.979} &  25.32 & \textbf{26.67} &    0.123 & \textbf{0.107} &  0.927 & \textbf{0.940}   \\
            cone      &  31.50 & \textbf{31.75} &  0.059 & \textbf{0.055} &  0.969 & \textbf{0.970} &   -- &   -- &  -- &  -- &  -- &  -- \\
            earphone  &  30.94 & \textbf{31.74} &  0.029 & \textbf{0.023} &  0.974 & \textbf{0.977} &  26.10 & \textbf{27.41} &    0.098 & \textbf{0.084} &  0.940 & \textbf{0.947} \\
            globe     &   \textbf{28.83} & 28.52 &  \textbf{0.039} & 0.043 & \textbf{0.958} & 0.955 &  \textbf{23.89} & 23.61 &    \textbf{0.129} & 0.131 &  \textbf{0.916} & 0.911 \\
            shoe      &  26.31 & \textbf{26.82} &  0.054 & \textbf{0.050} &  0.950 & \textbf{0.952} &  -- &   -- &  -- &  -- &  -- &  -- \\
            snowman   &  35.63 & \textbf{36.93} &  0.017 & \textbf{0.014} &  0.991 & \textbf{0.993} &  31.87 & \textbf{33.24} &    \textbf{0.049} & 0.050 &  0.977 & \textbf{0.978} \\
    \bottomrule
\end{tabular}
  \caption{Reconstruction quality of \textbf{Vanilla NeRF} \cite{mildenhall2020nerf} and \textbf{Plenoxels} \cite{plenoxels} trained via trajectories generated from \textbf{Ensemble} (Ours) against the \textbf{Random} baseline. Results here are based on a subset of 6 objects from our Objaverse test-set and averaged across a range of budgets. The best result is represented with \textbf{bold}.}
  \label{tab:result_3}
\end{table*}

\noindent{}\textbf{Primary Evaluation}. We first evaluate the performance of our model on the NeRF-Synthetic dataset with unseen classes and objects of higher complexity for a challenging out-of-distribution test. Note that \textit{chair} has been removed from the evaluation since our original training dataset contains said class. The result can be found in Tab. \ref{tab:result_2}. For this evaluation, we compare our performance against a prior approach, ActiveNeRF \cite{pan2022activenerf}. Additionally, we have also trained multiple instances of each object across a range of budget, with $\mathcal{B} \in \left\{3, 6, 12, 20, 30 \right\}$. This way, we should be able to get a better representation of the potency of each method. We reported the averaged result across said budgets on 3 reconstruction metrics of Peak Signal-to-Noise Ratio (PSNR), Structural Similarity Index Measure (SSIM) \cite{ssim} and Perceptual Similarity (LPIPS) \cite{lpips}.

Firstly, note that the reconstruction quality of a model trained via our Ensemble schema, \textbf{E} outperforms ActiveNeRF in almost every class and across all reconstruction quality metrics, lagging behind only on \textit{ship} by a small margin, demonstrating its potency. 
Secondly, since our proposed approach is model agnostic for radiance fields modeling, we can therefore decouple trajectory planning from the radiance fields training. Succinctly, this means, unlike \cite{pan2022activenerf, neurar} where they are locked into a particular radiance fields model, which could be very slow to train, our approach does not impose such a restriction and does not require any radiance fields training for trajectory planning. For these reasons, we are able to achieve a significant speed-up, with an average $\sim$80x improvement compared to the ActiveNeRF. In Tab. \ref{tab:result_2} we report the total time taken as two separate components - the \textit{trajectory generation} time, TG and the \textit{radiance fields training} time, RFT. For ActiveNeRF we report only a single value for both time components since the two processes are inherently one. Finally, we retrained additional TensoRF instances with the trajectories generated via ActiveNeRF with the same configurations so the reconstruction metrics would be comparable; for fairness, we have omitted the time taken for said retraining. It should be noted that our approach does not need the image taken from a candidate unseen pose for scoring, only the camera extrinsics, rendering it much more useful in real-world applications (e.g. does not need tedious sampling by actual visitation of poses). Renders of sample views of the methods can be found in Fig. \ref{fig:result_render} for visual comparison.

For the in-distribution evaluation, we trained TensoRF on the custom Objaverse final-set via 6 trajectory planning schemas - \textbf{Random} as a baseline, $\textbf{Ensemble}$ as described in Sec. \ref{subsec:method_trajectory_generation}, and the maximization of \textbf{individual objective functions} ${f_C}$, $f_Q$, $f_D$ and $f_D$. For the random baseline, we independently generated 5 trajectories per object and reported their averaged result. As mentioned prior, the optimal objective function could differ across objects based on their physical representation. As such, we have grouped the set of results trained on the trajectory generated via the \textit{individual} objective functions and took the best performing amongst them as another candidate schema dubbed, \textbf{Optimal CQDT}. The results can be found in Tab. \ref{tab:average_objaverse_metrics}; we used a simple baseline here due to limited computational resource.
\begin{table}
\centering
    \begin{tabular}{lccc}
        \toprule
        Method & \textbf{PSNR} ($\uparrow$) & \textbf{LPIPS} ($\downarrow$) & \textbf{SSIM} ($\uparrow$) \\
        \midrule
        Random & 29.07  & 0.081 & 0.951 \\
        E (Ours) & \underline{29.45}   & \underline{0.071}  & \underline{0.955}\\
        OP (Ours) & \textbf{29.67}  & \textbf{0.068}  & \textbf{0.957}\\
        \bottomrule
    \end{tabular}
  \caption{Averaged result of \textbf{Optimal CQDT, OP} and \textbf{Ensemble, E} against the \textbf{Random} baseline. The best result is represented with \textbf{bold}, and second best with \underline{underline}.}
  \vspace{-4mm}
  \label{tab:average_objaverse_metrics}
\end{table}

From the table, we can observe that Optimal CQDT performs the best, with Ensemble coming in as the second best, outperforming the Random baselines. The superior performance of Optimal CQDT reveals that the objects from the Objavserse dataset can be saliently captured and represented by a single objective function. However, it should also be noted that the efficacy may not be homogeneous across the objective functions for a given object. For example, for \textit{bookcase}, $f_D$ is worse than $f_C$. We hypothesized that this is due to the complexity of the shelving and large-flat surfaces, which creates a local minima for $f_D$, resulting in an over concentration around a local region. On the flip side, for \textit{soccer\_ball}, the random baseline is actually competitive compared to the other schemas. We hypothesized that for objects that are geometrically simple and easily covered from all directions, random walk could suffice since the difference in saliency across poses is minimal. From this observation, we further hypothesized that for such objects, one could simply utilize a cheap random walk for trajectory planning. The breakdown of the result by class can be found in the supplementary. 

Lastly, Ensemble does indeed offer a viable alternative to trajectory planning when the optimal objective function is not known ahead of time, or if it's too costly to be determined explicitly.
\\
\noindent{}\textbf{Model Variant Generalization}. Intuitively, our approach should also generalize across radiance field model variants since the objective functions are grounded in geometry and photometric visual cues. Therefore, we also tested for such generalization by evaluating the approach on a subset of our Objaverse dataset with a fully neural-implicit radiance fields model, Vanilla NeRF and a fully explicit model, Plenoxels. With these, our evaluations covered the spectrum of radiance fields models from fully neural-implicit to fully explicit models. The result of the evaluation can be found in Tab. \ref{tab:result_3}. For the results on \textit{cone} and \textit{shoe} with Plenoxels, we have omitted them as Plenoxels is unable to converge for them at all. The result validates our hypothesis by demonstrating that indeed, the proposed schema remain efficacious across model variants in almost all cases.



\section{Conclusion}\label{sec:conclusion}
\noindent{}\textbf{Limitation}. The current ensembling schema is quite simple and by no means exhaustive; our future studies will consider more effective ways to merge the proposed objective functions. Additionally, we currently focus only on inward-looking scenes with singular objects, which is not representative of the complex environment one may encounter in complex mapping and navigation tasks. Therefore, we would also like to apply SOAR to more challenging large-scale environments in future works.
\\

\noindent{}\textbf{Summary}. In this work, we have proposed SOAR, a set of objective functions designed to serve as surrogates to indicate the goodness of a resultant radiance fields model, given a set of inputs. In order to enable efficient inference for real-time trajectory generation, we further proposed a deep neural network, SOARNet that enables per-step planning in $<$1s on unseen poses. With extensive evaluations, we have shown that indeed our approach is able to outperform various baselines, even in challenging out-of-distribution scenarios and across radiance fields model variants.\\

\noindent{}\textbf{Acknowledgement} We extend our gratitude to several individuals and teams whose support contributed to the fruition of our work. We express our appreciation to Mohit Mehta and Anbang Yang for their helpful discussion, and to the NYU High Performance Computing team for their support in our experimentation. The work is supported by NSF Awards 2238968 and 2322242.




{
    \small
    \bibliographystyle{ieeenat_fullname}
    \bibliography{main}
}
\clearpage
\setcounter{page}{1}
\maketitlesupplementary

\section{Supplementary}
\label{sec:supplementary}
As supplementary, we provided more details of our proposed approach below. In Sec. \ref{subsec:supplementary_dataset}, we discuss about the dataset generation procedure used to generate our custom Objaverse \cite{deitke2023objaverse} based dataset. Then in Sec. \ref{subsec:supplementary_nerf_configs} we outlined the detailed configurations we used for the respective radiance fields model and baseline. Thereafter in Sec. \ref{subsec:supplementary_add_results}, a further drill-down of the results previously shown in Tab. 3 of Sec. 4.2 is provided. Finally, we provided a video visualizing the trajectories and resulting novel view synthesis renders of our method, SO-NeRF against ActiveNeRF \cite{pan2022activenerf} in Sec. \ref{subsec:supplementary_video_rendering}.

\subsection{Dataset Generation}
\label{subsec:supplementary_dataset}
\begin{figure}[t]
  \centering
    \includegraphics[width=0.85\linewidth]
    {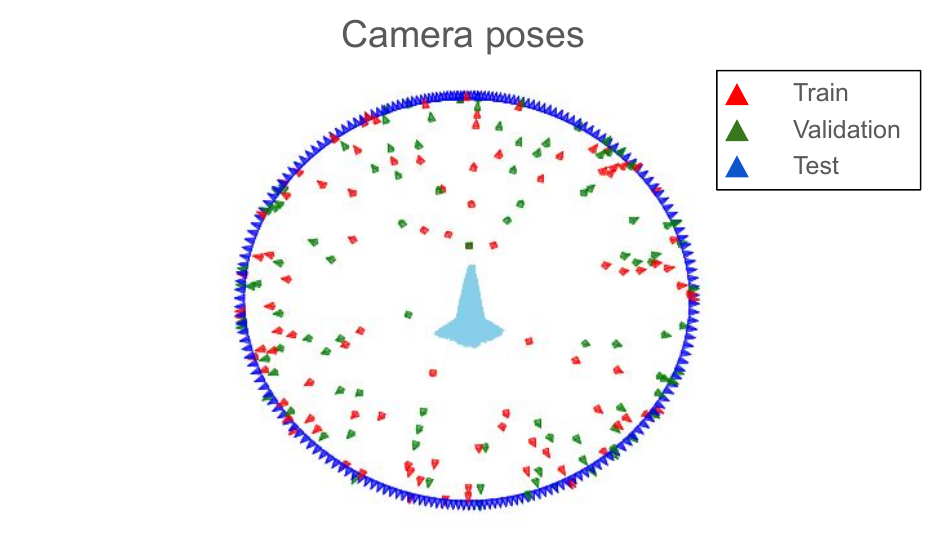}
    \vspace{-2mm}
    \caption{Visualization of the camera poses visited to sample views for the train, validation and test sets of the \textit{cone} object.}
    \vspace{-2mm}
    \label{fig:pose_preview}
\end{figure}
\begin{table*}[htbp]
\centering
    \begin{tabular}{lccc|ccc|ccc}
        \toprule
        {} & \multicolumn{3}{c}{\textbf{PSNR} ($\uparrow$)} & \multicolumn{3}{c}{\textbf{LPIPS} ($\downarrow$)} & \multicolumn{3}{c}{\textbf{SSIM} ($\uparrow$)} \\
        \midrule
        Class &  Rand &  E &  Op. CQDT &  Rand &  E &  Op. CQDT &  Rand &  E &  Op. CQDT \\
        \midrule
            bookcase    &        28.94 &        30.37 &         \textbf{30.42} &         0.089 &         0.078 &         \textbf{0.074} &        0.946 &        0.956 &         \textbf{0.957} \\
            chair       &        29.33 &        29.59 &         \textbf{30.00} &         0.084 &         0.069 &         \textbf{0.067} &        0.949 &        0.956 &         \textbf{0.957} \\
            cone        &        27.64 &        28.37 &         \textbf{28.59} &         0.100 &         0.088 &         \textbf{0.083} &        0.938 &        0.945 &         \textbf{0.947} \\
            control     &        29.33 &        30.41 &         \textbf{30.48} &         0.069 &         \textbf{0.055} &         \textbf{0.055} &        0.957 &        \textbf{0.965} &         \textbf{0.965} \\
            earphone    &        29.16 &        29.74 &         \textbf{30.05} &         0.074 &         0.060 &         \textbf{0.058} &        0.954 &        {0.961} &         {0.961} \\
            fan         &        \textbf{27.30} &        27.01 &         26.98 &         \textbf{0.091} &         0.092 &         0.093 &        \textbf{0.939} &        0.938 &         0.938 \\
            fireplug    &        30.21 &        30.37 &         \textbf{31.30} &         0.064 &         0.065 &         \textbf{0.054} &        0.959 &        0.959 &         \textbf{0.964} \\
            globe       &        25.54 &        26.05 &         \textbf{26.14} &         0.110 &         0.094 &         \textbf{0.089} &        0.915 &        0.925 &         \textbf{0.928} \\
            gravestone  &        31.43 &        31.52 &         \textbf{31.59} &         0.054 &         \textbf{0.050} &         0.051 &        0.967 &        0.968 &         \textbf{0.969} \\
            guitar      &        32.22 &        \textbf{32.43} &         32.33 &         0.053 &         \textbf{0.045} &         \textbf{0.045} &        0.976 &        \textbf{0.979} &         \textbf{0.979} \\
            lion        &        30.53 &        \textbf{30.70} &         30.57 &         0.063 &         0.053 &         \textbf{0.052} &        0.962 &        \textbf{0.965} &         \textbf{0.965} \\
            shoe        &        26.23 &        \textbf{26.97} &         26.73 &         0.099 &         \textbf{0.085} &         \textbf{0.085} &        0.931 &        \textbf{0.937} &         \textbf{0.937} \\
            snowman     &        32.01 &        32.49 &         \textbf{32.62} &         0.063 &         0.057 &         \textbf{0.056} &        0.966 &        0.970 &         \textbf{0.971} \\
            soccer\_ball &        \textbf{25.53} &        24.24 &         25.10 &         0.113 &         0.109 &         \textbf{0.097} &        0.941 &        0.939 &         \textbf{0.944} \\
            sunglasses  &        28.26 &        28.01 &         \textbf{28.61} &         0.093 &         0.079 &         \textbf{0.072} &        0.955 &        0.958 &         \textbf{0.962} \\
            teddy\_bear  &        30.92 &        31.44 &         \textbf{31.53} &         0.064 &         0.055 &         \textbf{0.054} &        0.960 &        \textbf{0.965} &         \textbf{0.965} \\
            toilet      &        28.95 &        29.53 &         \textbf{30.13} &         0.098 &         0.086 &         \textbf{0.077} &        0.943 &        0.950 &         \textbf{0.954} \\
            vase        &        29.75 &        30.92 &         \textbf{30.97} &         0.073 &         \textbf{0.063} &         0.064 &        0.952 &        \textbf{0.959} &         \textbf{0.959} \\
        \bottomrule
    \end{tabular}
  \caption{Reconstruction quality of \textbf{TensoRF} trained via trajectory generated from \textbf{Optimal CQDT} and \textbf{Ensemble, E} against the \textbf{Random} baseline. Results here are grouped by object class and averaged across the budgets. The best result is represented with \textbf{bold}.}
  \label{tab:result_1}
\end{table*}

 To begin, we generated our custom dataset using the Objaverse API \cite{deitke2023objaverse}. Given the large variety of classes and object instances per class that are available within the Objaverse dataset, there is an expected amount of noise to the labels. For this reason, we specifically select only Large Vocabulary Instance Segmentation \cite{gupta2019lvis} (LVIS) objects within Objaverse, which have been manually verified by humans.

Just like the NeRF Synthetic dataset, our custom data generation procedure samples views from poses around the object constrained to a fixed radius; this is equivalent to sampling views from the surface of an imaginary sphere. For the training and validation set, we performed the sampling randomly around the object, whereas for the test set, we used a smooth circular trajectory instead for ease of viewing. An illustration of the aforementioned sampling procedures can be found in Fig. \ref{fig:pose_preview}. The procedure is summarized in Algorithm \ref{algo:processglb}. We ran the algorithm on a machine with a single AMD Radeon MI50 GPU, $4$ CPUs and 16GB of DRAM.

\begin{figure*}[t]
  \centering
    \includegraphics[width=0.99\linewidth]
    {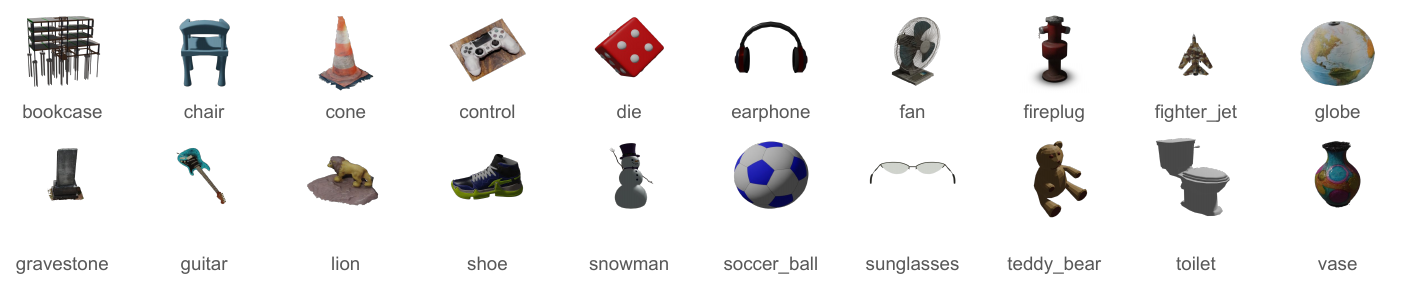}
    \vspace{-2mm}
    \caption{Preview of an object from each class of our custom Objaverse dataset used to train and evaluate SOARNet.}
    \vspace{-2mm}
    \label{fig:objaverse_preview}
\end{figure*}

Overall, we sampled 640 objects across $20$ classes - specifically,  \textit{bookcase}, \textit{chair}, \textit{cone}, \textit{control}, \textit{die}, \textit{earphone}, \textit{fan}, \textit{fireplug}, \textit{fighter\_jet}, \textit{globe}, \textit{gravestone}, \textit{guitar}, \textit{lion}, \textit{shoe}, \textit{snowman}, \textit{soccer\_ball}, \textit{sunglasses}, \textit{teddybear}, \textit{toilet}, and \textit{vase} from the list of available categories. For each class, we downloaded twenty instances of that class and manually verified the labels for reliability. We pruned or replaced any outlier objects. Visualizations of the object classes can be found in Fig. \ref{fig:objaverse_preview}.


\subsection{Additional Model Configuration Details}
\label{subsec:supplementary_nerf_configs}

For radiance fields modeling, we leveraged the Nerfstudio \cite{nerfstudio} implementations of TensoRF \cite{Chen2022ECCV} and Vanilla NeRF \cite{mildenhall2020nerf} for training. For TensoRF, we trained each object-instance for $300$ epochs with all other configurations left as default. For Vanilla NeRF, we trained each object-instance for 3,000 epochs instead for better convergence, with all other configurations left as default as well.
\begin{algorithm}[t]
\caption{ProcessGLB}\label{algo:processglb}
\begin{algorithmic}[1]
\Procedure{ProcessGLB}{}
    \State $\Call{CleanScene}$ \Comment{Clean the blender scene}
    \State $obj\gets\Call{LoadObject}{glb\_file\_path}$
    \State $\Call{ScaleObject}{obj}$ \Comment{Normalize scale}
    \State $imgs,cam\_parameters\gets\Call{SampleViews}$
    \State $\Call{SaveObj}{obj}$ \Comment{Export obj + mtl file}
    \State $\Call{SaveSamples}{imgs,cam\_parameters}$
\EndProcedure
\end{algorithmic}
\end{algorithm}
For Plenoxels \cite{plenoxels}, we trained it with an epoch-size of 300 per object, with a learning rate of 3e1, and spherical harmonics learning rate of 1e-2 with all other configurations as default.  In terms of hardware, both models were trained on a platform with an NVIDIA RTX8000 GPU, 32GB of DRAM, and $2$ CPUs.

For our ActiveNeRF baseline, all the trajectory planning and generation runs were conducted on a machine with an NVIDIA RTX8000 GPU, 64GB of system DRAM, and 8 CPUs for training, per object. We initialized each instance with a seed training set with $5$ poses and let ActiveNeRF determine the required remaining images based on the quantified uncertainty. Closely following the experimental configuration of \cite{pan2022activenerf}, we trained the model for 200,000 iterations each, and let it pick the remaining $B-5$ number of images over the course of those iterations at a cadence of 5 images per 40,000 steps and minimum uncertainty kept at 0.01. The remaining settings were all kept at default.

\subsection{Additional Result Breakdown}
\label{subsec:supplementary_add_results}
In Sec. 4.2, the averaged results on the Objaverse dataset were reported in Tab. 3. In this section, we go one step further by breaking down the result per-class to provide better clarity on the performance of our proposed approach - \textbf{Ensemble} and \textbf{Optimal CQDT} as previously defined - against the \textbf{Random} baseline in Tab. \ref{tab:result_1}.

As we can observe from the table, the proposed approaches worked the best in almost all cases, except for \textit{soccer\_ball} and \textit{fan}, across all 3 metrics. The results shown here affirm the findings and conclusion we put forth previously.

\subsection{Video Rendering}
\label{subsec:supplementary_video_rendering}
For additional visualization and demonstration of our approach, we have included a short video together with the supplementary submission to illustrate the trajectory generated and test-set renderings of SO-NeRF against ActiveNeRF on the NeRF Synthetic dataset. 



\end{document}